# Analysis, Identification and Prediction of Parkinson's Disease Sub-Types and Progression through Machine Learning

**Ashwin Ram**

Department of Electrical and Computer Engineering, University of Texas at Austin, Austin, USA
Email: ashwin.ram@utexas.edu

## Abstract

This paper represents a groundbreaking advancement in Parkinson's disease (PD) research by employing a novel machine learning framework to categorize PD into distinct subtypes and predict its progression. Utilizing a comprehensive dataset encompassing both clinical and neurological parameters, the research applies advanced supervised and unsupervised learning techniques. This innovative approach enables the identification of subtle, yet critical, patterns in PD manifestation, which traditional methodologies often miss. Significantly, this research offers a path toward personalized treatment strategies, marking a major stride in the precision medicine domain and showcasing the transformative potential of integrating machine learning into medical research.

## 1. Introduction and Related Work

As stated by Pringsheim *et al.* (2014), Parkinson's disease (PD) is among the most frequently encountered neurodegenerative disorders, present in about four percent among individuals aged eighty and above [1]. Although the reduction of dopaminergic neurons in the midbrain is widely recognized as a significant fac-tor in contributing to Parkinson's disease (PD), the majority of cases still have an elusive etiology. Consequently, there remains a profound lack of understanding regarding the diverse patterns of disease progression observed among patients. In response to this issue, since 2010, the Parkinson's Progression Markers Initiative (PPMI) have been actively gathering rich longitudinal data from distinct patient groups and utilizing various data modalities in an effort to tackle this issue, encompassing a diverse range of clinical measurements including neuroimaging scans, expression profiles for genes, protein levels, data captured from sensors and wearable devices, and genomic variant statuses (Marek *et al.*, 2011, 2018) [2] [3]. This extensive dataset has a primary aim of identifying noteworthy biomarkers that can facilitate the development of innovative interventions for Parkinson's disease. With its comprehensive and meticulously annotated dataset, therefore, the PPMI initiative facilitates the investigation of various biological variables in correlation with clinical markers of disease severity, allowing for a thorough exploration of Parkinson's disease.

There have been over 110 machine learning studies that used the PPMI database, out of which 97 employed supervised learning, compared to only 19 that employed an unsupervised learning approach. Within the subset of studies that reported supervised learning models in their research, 55 supervised-based studies were dedicated to predicting Parkinson's diagnosis, indicating both their prevalence and significance in the research. While early detection of Parkinson's disease holds significance, it is worth noting that established clinical tools for diagnosing PD already exist [4]. Hence, machine learning approaches centered around diagnosis are unlikely to significantly contribute to the primary objective of the PPMI study, which aims to comprehend the diverse symptomatology of patients and their progression patterns over time. Only 26 Machine Learning studies utilized the longitudinal structure of PPMI data to anticipate future symptoms based on a starting point, referred to as "progression prediction." As the PPMI project aims to comprehend the biological factors linked to variations in patient trajectories, these progression prediction papers hold significant importance. Thirteen more studies concentrated on predicting symptoms measured concurrently with the predictive features, while five studies' foci were on neuroimaging outcomes or medication status rather than symptoms or diagnosis.

A limited number of studies employed unsupervised learning techniques to generate latent variables or clusters, aiming to capture the variability among patients. Out of the 19 studies that applied unsupervised techniques, eleven focused on subtyping Parkinson's patients through clustering models. Additionally, eleven studies employed methods involving latent variables or dimensionality reduction, utilizing continuous latent factors. Furthermore, three studies utilized both subtyping and continuous latent variables in their analyses. Remarkably, a mere six papers managed to amalgamate supervised and unsupervised methodologies, despite the proclaimed emphasis in PD research on identifying sub-types capable of forecasting distinct progression patterns among various groups of PD patients. In order to effectively detect underlying categories among patients and reveal factors that can predict their future category memberships, it is highly likely that a combination of supervised and unsupervised models will be required. Noteworthy is the fact that three papers successfully integrated patient clustering into subtypes with the prediction of existing or forthcoming symptoms. For example, Faghri *et al.* (2018) employed a combined approach of Non-negative Matrix Factorization (NMF) and Gaussian Mixture Models (GMMs) to cluster patients into subtypes [5]. They then utilized random forests for supervised prediction of symptom levels four years later [5]. In a similar vein, Valmarska *et al.* (2018) utilized unsupervised clustering techniques to categorize patients based on the Movement Disorder Society Unified Parkinson's Disease Rating Scale (MDS-UPDRS) part III scores [6]. They further developed a supervised algorithm to determine the predictive features influencing changes in cluster assignment over time, with bradykinesia emerging as the most influential attribute within their model. In another study, Zhang *et al.* (2019) employed Long Short-Term Memory (LSTM) networks to encode sequences of clinical observations [7]. They utilized Dynamic Time Warping (DTW) to estimate the similarity between LSTM activation sequences for each pair of patients. To condense patient data into a two-dimensional representation while maintaining the integrity of DTW distances, the researchers utilized the Student t-distributed Stochastic Neighbor Embedding (t-SNE) technique. Afterward, the patients were categorized into three distinct subtypes using k-means clustering within this condensed space. Moving forward, it is imperative for future research to focus on integrating supervised and unsupervised models. This integration will enable the exploration of subtypes or other latent variables that can elucidate the heterogeneity in patient characteristics while simultaneously predicting future outcomes.

This research paper employed a methodology that addressed the lack of previous studies, integrating the clustering of patients into subtypes with the prediction of current or future symptoms: we combine supervised and unsupervised machine learning methods in order to identify subtypes that can accurately predict progression across distinct groups of Parkinson's patients. We amalgamate unsupervised patient clustering into subtypes with the ability to predict their present or future symptoms, and we work modern longitudinal data with different data dictionaries and data labels given in the Progressive Parkinson's Markers Initiative (PPMI) database. In addition, we were able to perform accurate prediction at baseline: namely, given a PD patient today, we were able to determine their disease trajectories and symptoms in an accurate manner immediately.

## 2. Methodology

The methodology employed in this study is structured into distinct phases, as

outlined below:

1) **Data Preprocessing:** Initial data preprocessing involved normalization using the min-max method followed by longitudinal data vectorization.

2) **Dimensionality Reduction:** We created a Parkinson's Progression Space by applying Non-negative Matrix Factorization (NMF) and analyzing the resulting latent vectors.

3) **Unsupervised Clustering:** The subtypes of Parkinson's Disease were identified and clustered using an unsupervised Gaussian Mixture Model (GMM) approach.

4) **Model Replication:** The GMM was then replicated on the Parkinson's Disease Biomarker Discovery (PDBP) database to validate the clustering patterns across distinct datasets.

5) **Supervised Learning:** With the unsupervised approach established, we progressed to employing supervised learning techniques, specifically ensemble methods such as Random Forest Classification, to predict Parkinson's disease subtypes at baseline.

6) **Supervised Model Utilization:** The supervised model, trained on the PPMI database, was utilized to predict disease subtypes within the PDBP database.

7) **Validation:** We concluded with a 5-fold cross-validation of the model results to ensure robustness and reliability.

## 2.1. Data Preprocessing

The data processing pipeline was meticulously structured and involved the following steps:

1) **Data Acquisition:** We aggregated a comprehensive dataset comprising various patient assessments and tests, including Family History, Motor and Non-Motor Skills, Biospecimen Analysis, and an array of cognitive and behavioral evaluations such as the Montreal Cognitive Assessment, Hopkins Verbal Learning Test, and REM Sleep Behaviour Disorder Questionnaire. The dataset spanned across different patient groups: $n = 450$ Parkinson's Disease (PD) patients, Prodromal patients, Healthy Controls (HC), and individuals diagnosed with Parkinson's but showing normal imaging results (SWEDD).

2) **Data Vectorization:** Time-series data from all enrollment patients was transformed into a unified series using vectorization:

$$\mathbf{V} = \mathrm{Vec}\left(\mathbf{X}_1, \mathbf{X}_2, \cdots, \mathbf{X}_n\right) \tag{1}$$

where $\mathbf{V}$ represents the vectorized time-series and $\mathbf{X}_i$ represents the individual time-series data for each patient. Non-progression-related data were omitted in this step.

3) **Normalization:** All vectorized time-series data $\mathbf{V}$ were subjected to normalization. We employed two methods: z-score and min-max normalizations, defined as:

$$V'_{\text{z-score}} = \frac{V - \mu}{\sigma} \tag{2}$$

$$V'_{\text{min-max}} = \frac{V - \min(V)}{\max(V) - \min(V)} \tag{3}$$

where $V'$ is the normalized vector, $\mu$ is the mean of the vector $V$, and $\sigma$ is the standard deviation of $V$. Our evaluation confirmed that the min-max normalization preserved the progression pattern within the dataset.

#### 2.1.1. Vectorization of Time-Series Data

All time-series data from these tests were transformed into a unified series. This process is represented as:

$$\mathbf{V} = \text{vectorize}(\text{CNE}, \text{MoCA}, \text{HVLT}, \text{ESS}, \text{SFT}, \cdots)$$

where $\mathbf{V}$ represents the vectorized data.

#### 2.1.2. Normalization Techniques

The normalized data is crucial for consistent analysis. Two normalization methods were evaluated:

1) Z-score normalization, defined as:

$$Z = \frac{X - \mu}{\sigma}$$

where $X$ is the original data point, $\mu$ the mean, and $\sigma$ the standard deviation.

2) Min-max normalization, maintaining the progression pattern:

$$X_{\text{norm}} = \frac{X - X_{\min}}{X_{\max} - X_{\min}}$$

where $X_{\max}$ and $X_{\min}$ are the maximum and minimum values in the data set, respectively.

### 2.2. Dimensionality Reduction to Create Parkinson's Progression Space

We then employed dimensionality reduction in order to make intuitive sense of the longitudinal data in order to build a progression space that allows for approximations of a given PD patient's trajectory given their relative location in that space. Albeit we used Principal Component Analysis (PCA), Non-Negative Matrix Factorization (NMF) and Independent Component Analysis (ICA) methods to perform the dimensionality reduction, the NMF worked best by mapping mathematically linked parameters onto a multi-dimensional space, which results in the proximity of comparable data points, thereby helping collapse the parameters. Specifically, one matrix comprises of latent vectors representing the progression space, while the second matrix includes indicators for progression stands associated with the latent vectors, serving as a linkage between the symbolic and real-world data. Through a thorough examination of the matrix containing latent vectors in the progression space, we can unveil the associated mapping and acquire valuable insights into the symbolic dimensions of the modeled progression space. With the aid of Nonnegative Matrix Factorization (NMF), we have effectively identified the primary symptom patterns in primary progressive diseases, encompassing motor impairments, cognitive dysfunction, and disturbances related to sleep.

The dimensionality reduction is achieved using a combination of Principal Component Analysis (PCA), Independent Component Analysis (ICA), and NMF. Formally, NMF is chosen due to its superior performance and is defined as:

$$V \approx W \cdot H$$

where $V$ represents the original data matrix, $W$ the basis matrix containing latent vectors, and $H$ the coefficient matrix.

The basis matrix $W$ in NMF is utilized to map the progression space of PD, which aligns closely with patient data. The coefficient matrix $H$ then links these latent vectors to progression indicators. The relationship is mathematically modeled as:

$$W_{i,j} \rightarrow \text{Latent vector for progression space}$$

$$H_{j,k} \rightarrow \text{Progression indicator for patient } k$$

where $i, j, k$ are indices representing specific features, latent vectors, and patients, respectively.

#### 2.2.1. Analytical Outcomes

Through NMF, we achieve a multi-dimensional representation of PD progression, capturing essential aspects like motor impairments, cognitive dysfunctions, and sleep disturbances. The analysis involves examining the matrix $W$ to understand the progression space, and $H$ for patient-specific progression patterns.

#### 2.2.2. Symbolic Representation of Progression Space

The symbolic dimensions in the modeled progression space are uncovered by analyzing $W$, providing insights into the nonlinear progression of PD symptoms. This analysis helps in mapping the complex trajectory of PD, expressed as:

$$\text{Symptom Pattern} = f\left(\text{Latent vectors in } W\right)$$

where $f$ represents the mapping function derived from NMF analysis.

### 2.3. Identifying and Clustering Parkinson's Subtypes

In order to perform the clustering, we used the unsupervised Gaussian Mixture Model (GMM). By utilizing GMM, the data was able to cluster itself into distinct groups based on the declination rate across numerous symptoms: from PD subtypes to non-PD controls (members of the control group). The reason why GMM is so powerful is that it captures natural distributions by assuming that the data is produced from a mixture of independent and identically distributed Gaussian probability distributions.

Through the utilization of GMM, we can effectively perform PD Progression projection, enabling a comprehensive analysis that explores the normalized projection trajectories of each sample relative to others, based on their classification. The progression velocity encompasses three key dimensions: motor impairments, cognitive decline, and sleep-related disturbances. Upon examining the cluster projected by the Parkinson's progression space, we observe that the motor dimension exhibits the highest variance, followed by sleep-related disturbances, and finally cognitive impairment. Within these trajectories, this learning approach categorizes PD patients into three distinct subtypes, aligning with the pace at which the disease progresses. Specifically, individuals with a slow progression rate are identified as PDVec1, while those with a moderate progression rate are labeled as PDVec2, and those experiencing rapid progression are denoted as PDVec3.

## 2.4. Supervised Machine Learning for Predicting Parkinson's Subtypes at Baseline

Expanding our investigation beyond the PPMI cohort, we employed GMM on the PPMI data, allowing us to encompass the PDBP cohort, characterized by a distinct recruitment strategy and design. Our discoveries unveil that the identified subtypes within the PDBP cohort display a comparable progression pattern to those observed within the PPMI cohort, indicating the consistency and generalizability of the model across diverse databases of Parkinson's disease patients.

Once the different progression classes were stratified and corroborated, we then performed supervised predictions that essentially determine the overall symptom of Parkinson's after not only 48 months but after 24 months, 12 months, and immediately. Compared to other supervised ensemble methods such as LASSO-regression or SVMs, the Random Forest model (RF) performed the best. Moreover, the RF model is stronger for the following trifecta of ideas: RF will determine the probability distribution of belong to a specific class which is key in our case because we want to track progression on an individual level; RF can handle a mixture of features (be them categorical or numerical); RF can naturally rank variables in a nuanced, meaningful way, quite significant for a classification problem. Furthermore, from this process, we then created models with varying levels of input factors, (baseline, baseline and first year, baseline followed by next two years), in order to predict the corresponding category (ie: progression class) of a given individual at a particular time period after the training. To validate our findings, we implemented two distinct validation methods. Initially, we conducted a comparable examination on an independent PDBP cohort to assess the performance metrics of the model. Additionally, we employed five-fold cross-validation on the PPMI dataset. This involved dividing the dataset into five random subsamples, where one subsample was designated as the validation data while the remaining four subsamples served as training data. We repeated this process five times, ensuring that each fold was compre-hensively covered.

## 3. Results

Figure 1 and Figure 2 show the Visualization of PD Progression space in 2D based on Dimension Reduction using 3 different techniques, namely PCA, Non-Negative Matrix Factorization (NMF) and FastICA techniques. NMF performed very well compared to PCA and FastICA techniques due to the non-negative nature of the clinical test results from PPMI. This process collapses mathematically related parameters into the same multi-dimensional space, mapping similar data points close together.

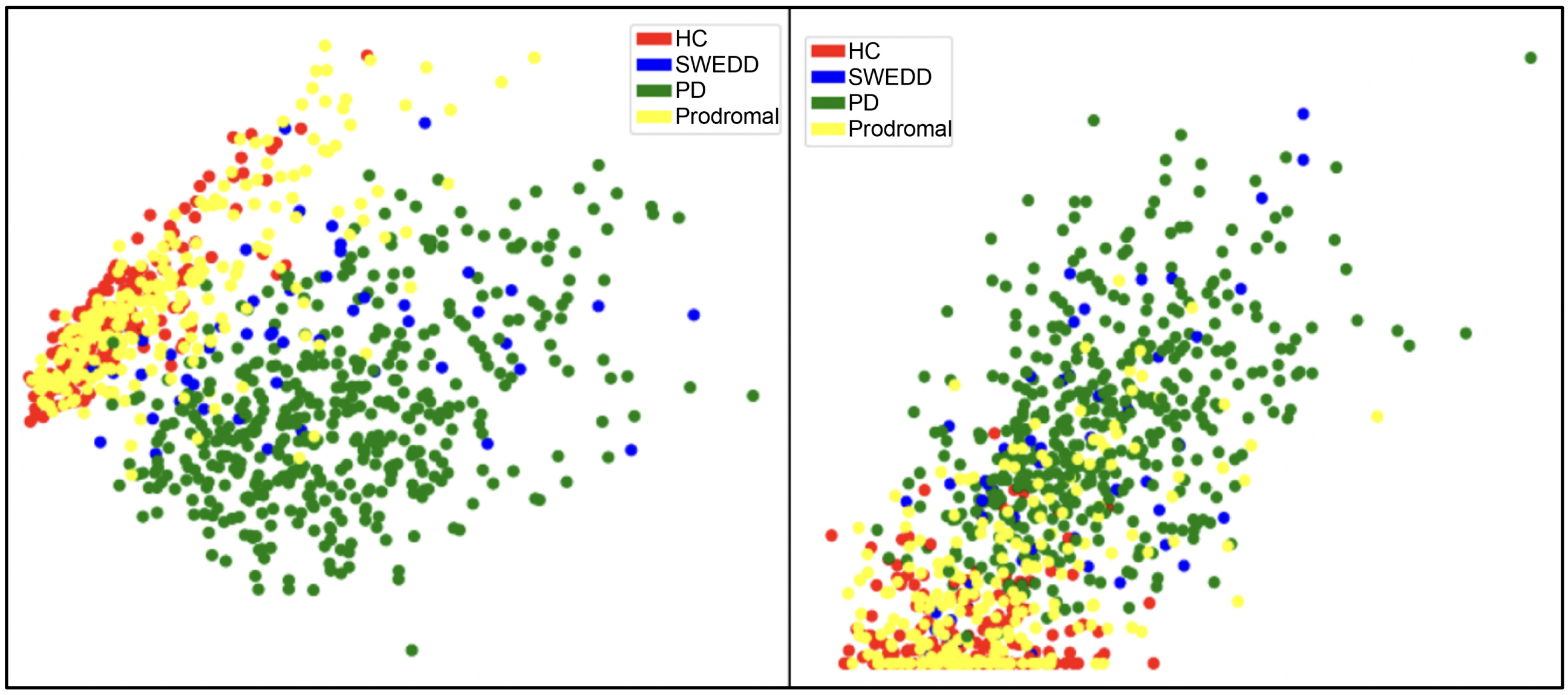

Figure 1. Left. Dimension Reduction using PCA. Right: Dimension Reduction using NMF.

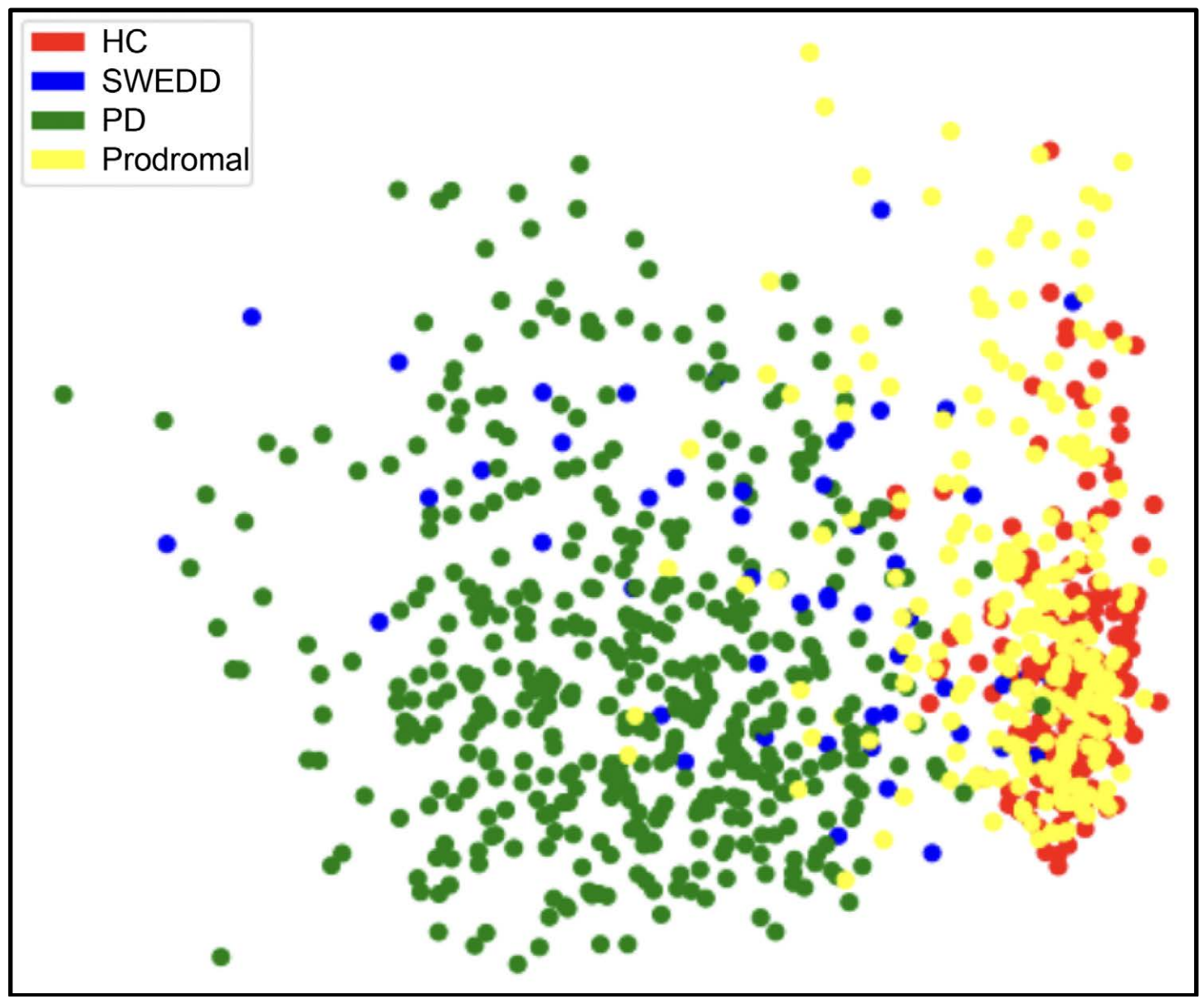

Figure 2. Dimension reduction using FastICA.

### 3.1. Principal Component Analysis (PCA)

The left plot in Figure 1 illustrates the result of PCA. The variance along the first two principal components shows a noticeable spread of data points, which suggests that PCA can effectively capture the dataset's variability. Different clusters are observable, indicated by the color coding, although some overlap between classes is evident. This overlap may imply that while PCA reduces dimensionality, it might preserve some relationships between classes that are not distinctly separable in the first two principal components.

### 3.2. Non-Negative Matrix Factorization (NMF)

The right plot in Figure 1 depicts the outcome of NMF. The data points are spread distinctly, with some degree of separation between classes. The non-negativity constraint of NMF leads to a parts-based representation, which in our case appears to provide an advantage in class discrimination. The clusters, represented by colors, show less overlap compared to PCA, indicating that NMF may reveal a more interpretable and separable structure within the data for this particular dataset.

### 3.3. Comparative Analysis

Comparing the two plots, NMF shows a potential for better class separation than PCA. This could be due to the additive-only combinations allowed by NMF, which accentuate features unique to each class. In contrast, PCA's linear combinations, which include both additive and subtractive aspects, might dilute these unique features. Therefore, for datasets where interpretability and parts-based representation are crucial, NMF might be the preferred method over PCA.

### 3.4. Fast Independent Component Analysis (FastICA)

Figure 2 demonstrates the dimensionality reduction using FastICA, a technique that identifies independent components within the data. In the context of Parkinson's disease progression, the color-coded scatter plot exhibits distinct groupings corresponding to different stages of the disease: Healthy Controls (HC), Scans without Evidence of Dopaminergic Deficit (SWEDD), Parkinson's Disease (PD), and the Prodromal phase.

#### 3.4.1. Observations

The data points show a discernible gradient from HC to PD, suggesting a potential trajectory of disease progression. Notably, the Prodromal stage points are predominantly situated nearer to the HC cluster, indicating a closer similarity to the healthy state than to the advanced PD stage. Conversely, the SWEDD points are interspersed between the HC and PD groups, hinting at the heterogeneity within this category and the possible overlap in disease manifestation.

#### 3.4.2. Implications

The ability of FastICA to segregate these stages into distinct clusters provides valuable insights into the progression of Parkinson's disease. The separation between the stages observed in the plot could have significant implications for early diagnosis and the understanding of disease mechanisms. The positioning of SWEDD and Prodromal points relative to HC and PD may also reflect varying degrees of neurodegeneration or compensatory mechanisms at play during the disease's early and intermediate stages.

#### 3.4.3. Analytical Significance

The visualization facilitated by FastICA underscores the potential of independent component analysis in biomedical research, particularly in disorders with progressive pathology such as Parkinson's disease. The clear demarcation of disease stages supports the utility of FastICA in exploring complex biological datasets where uncovering underlying independent factors is crucial for disease characterization and stratification.

### 3.5. 3D Visualization of Parkinson's Disease Progression

Figure 3 presents a three-dimensional scatter plot generated using Non-negative Matrix Factorization (NMF) to visualize the progression space of Parkinson's disease (PD). The axes correspond to different symptomatic dimensions: cognitive impairment, movement disorders, and sleep disorders.

#### 3.5.1. Analysis

The PD patients, represented by green points, are predominantly located within a specific region of the plot, suggesting a commonality in symptom manifestation. In contrast, Healthy Controls, depicted in red, are tightly clustered, indicating minimal symptomatic expression. This visual separation highlights the effectiveness of NMF in distinguishing between the affected and healthy subjects based on the three symptom dimensions.

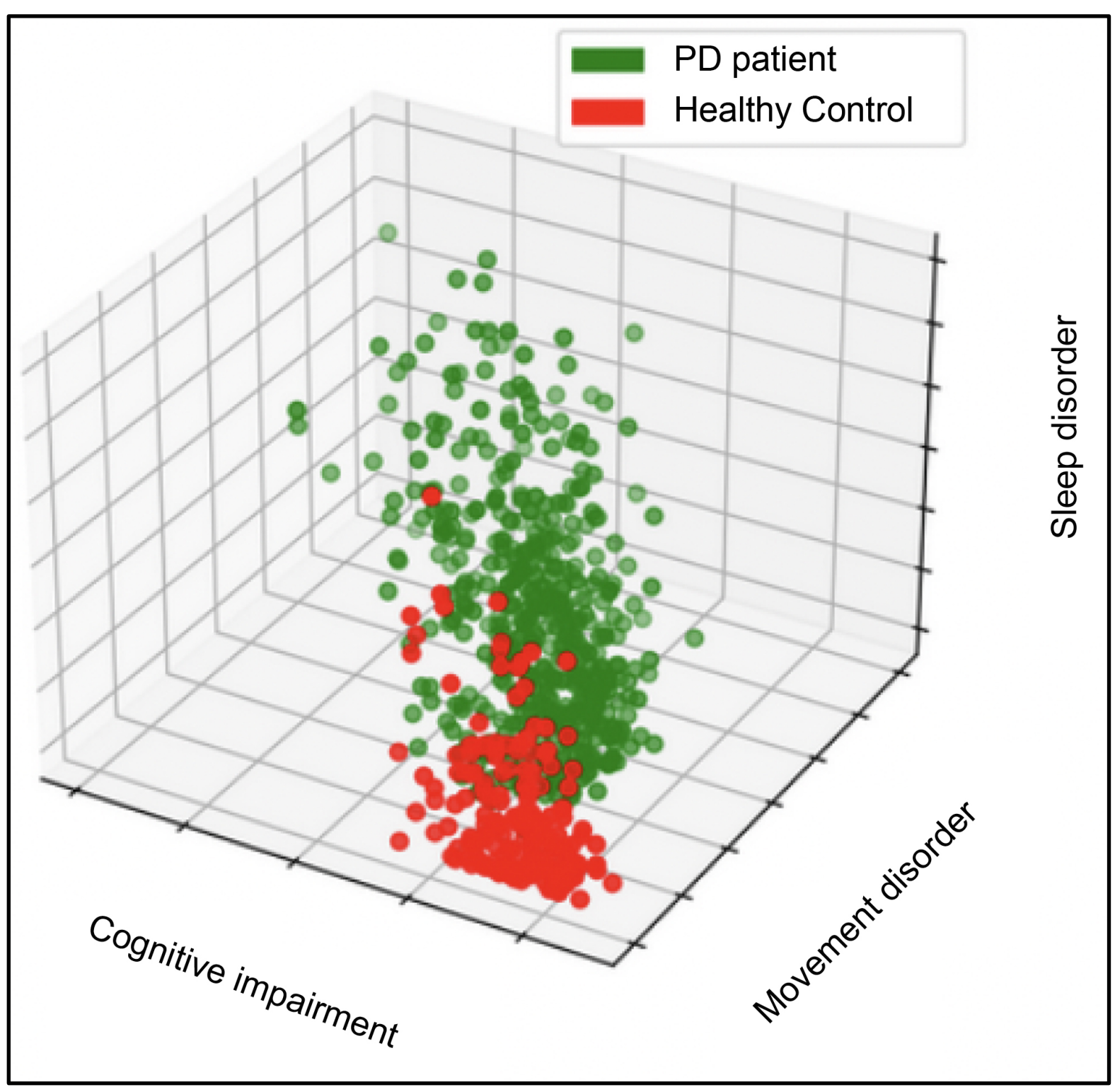

Figure 3. Generating a 3D visualization of the progression space using NMF.

#### 3.5.2. Interpretation

The spatial distribution of PD patients across the axes suggests variability in symptom severity and combination, which is consistent with the heterogeneous nature of PD. The distance between the two clusters may reflect the degree of deviation from normal health conditions, providing a potential measure for the severity of PD progression.

#### 3.5.3. Conclusion

The clear demarcation between PD patients and Healthy Controls in this 3D space underscores the potential of NMF in capturing the complex interplay of symptoms that characterize PD, offering insights into its progression and possibly aiding in the development of targeted interventions.

As indicated below, the figure exhibits a graphical depiction of unsupervised learning employing Gaussian Mixture Model (GMM) within a 2D progression space. The motor component is depicted along the x-axis, while the combined cognitive and sleep components are represented along the y-axis. These projected dimensions have been normalized, indicating that higher values in either direction signify a more pronounced decline. By employing GMM, the data is segmented into distinct subtypes that correspond to the rate of decline across various symptoms in comparison to non-PD controls. Through the use of the Bayesian information criterion, three Gaussian distributions have been identified, each representing a specific PD subtype. These subgroups, determined algorithmically within the case population, exhibit diverse patterns of change over time within the progression space and across specific biomarkers of progression. Notably, PDvec3 demonstrates a notably steeper progression slope.

### 3.6. Gaussian Mixture Model (GMM) Visualization

Figure 4 represents the visualization of a Gaussian Mixture Model (GMM) in a two-dimensional progression space relevant to Parkinson's disease. GMM is a probabilistic model that assumes all the data points are generated from a mixture of a finite number of Gaussian distributions with unknown parameters.

#### 3.6.1. Mathematical Description

The GMM is mathematically represented as:

$$p(\mathbf{x}) = \sum_{i=1}^{K} \phi_i \mathcal{N}(\mathbf{x} \mid \boldsymbol{\mu}_i, \boldsymbol{\Sigma}_i) \tag{4}$$

where $\mathbf{x}$ is a data point in the progression space, $K$ is the number of Gaussian components, $\phi_i$ are the mixing coefficients, and $\mathcal{N}(\mathbf{x} \mid \boldsymbol{\mu}_i, \boldsymbol{\Sigma}_i)$ are the component Gaussian densities, each with its own mean $\boldsymbol{\mu}_i$ and covariance $\boldsymbol{\Sigma}_i$.

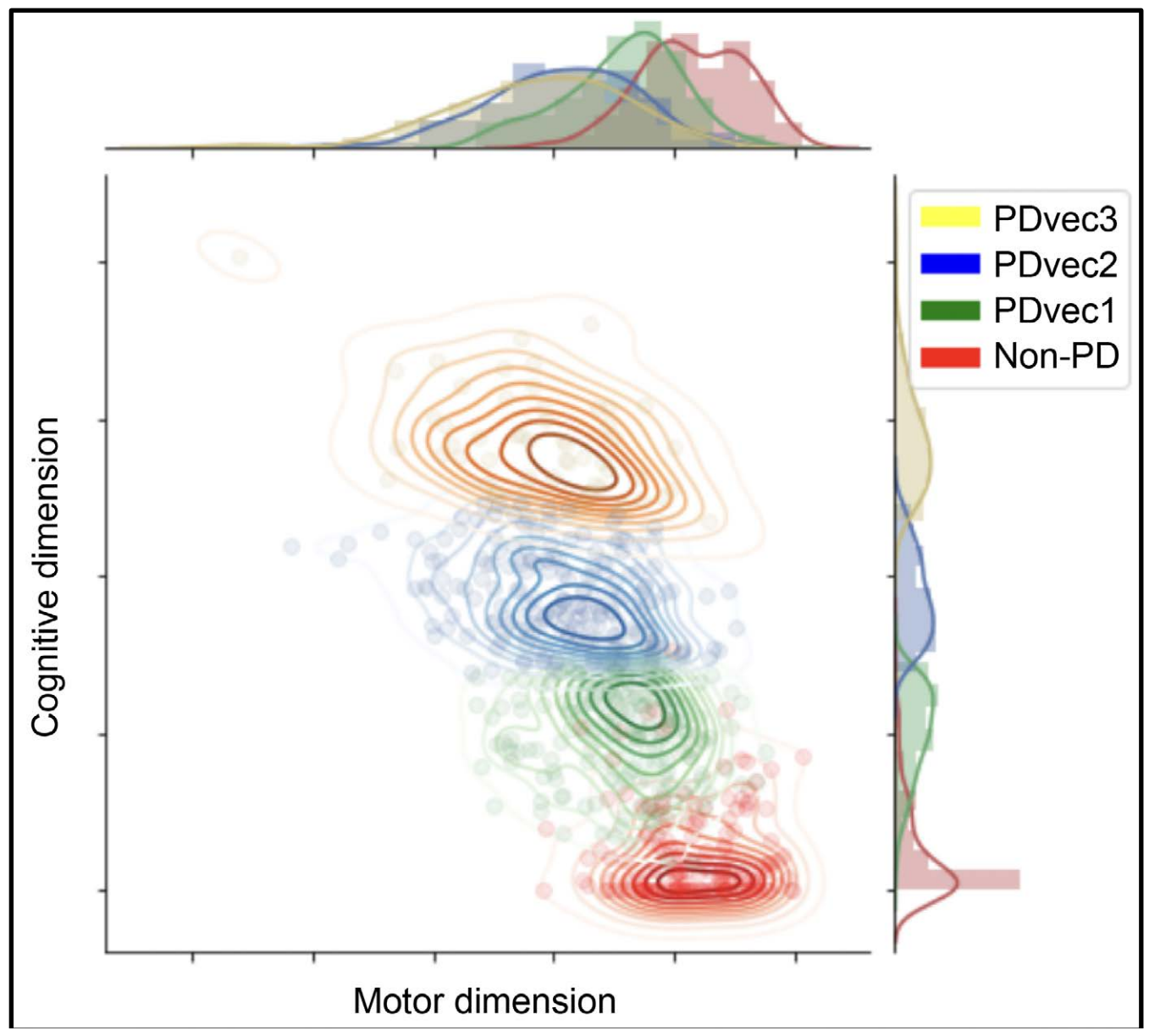

**Figure 4.** Visualization of GMM in a 2-dimensional progression space.

#### 3.6.2. Analysis of Parkinson's Progression

The figure illustrates the distribution of disease progression in terms of cognitive and motor dimensions, categorized into different severity stages: PDvec1, PDvec2, PDvec3, and Non-PD. The contour lines represent the iso-density lines of the Gaussian components, providing a visual representation of the density of data points at different stages of Parkinson's disease.

#### 3.6.3. Observations

1) PDvec1, PDvec2, and PDvec3 likely correspond to increasing severity of Parkinson's disease symptoms.

2) The overlap between PDvec1 and Non-PD contours suggests that early stages of Parkinson's might be difficult to distinguish from non-pathological aging.

3) PDvec3 shows a higher density in the region with severe motor and cognitive symptoms, indicating advanced Parkinson's.

#### 3.6.4. Implications

The GMM visualization in the context of Parkinson's disease can aid in understanding the overlap and distinction between various stages of the disease, potentially improving diagnosis and classification of the disease severity.

#### 3.6.5. Conclusion

GMM serves as a powerful tool for modeling the progression of Parkinson's disease, capturing the heterogeneity and overlap in symptomatology. The clarity in separation between the different stages of Parkinson's as shown in the figure reinforces the potential of GMM in clinical decision-making and personalized medicine.

As shown in the below **Figure 5**, we examine the dispersion of projected di-mensions, namely cognitive, motor, and sleep, among various categories of Parkinson's disease patients and healthy controls. The motor and sleep dimensions demonstrate an increase in disruptions, while the cognitive dimension indicates a decline. Notably, PDVec1 showcases the most pronounced levels of motor and sleep disturbances, alongside cognitive deterioration.

Figure 6 exhibits the evaluation of the Parkinson's disease Progression Model's performance. The receiver operating characteristic (ROC) curve demonstrates the predictive model's efficacy at the baseline stage, which was constructed using the PPMI cohort and assessed through five-fold cross-validation. This model correctly distinguishes patients with PD based on baseline only input factors and predicts their prognosis with an average AUC of 0.886 (0.89 for PDVec1, 0.84 for PDVec2, 0.93 for PDVc 3). The enhanced accuracy of PDVec3 can be attributed to the greater availability of subject-specific information.

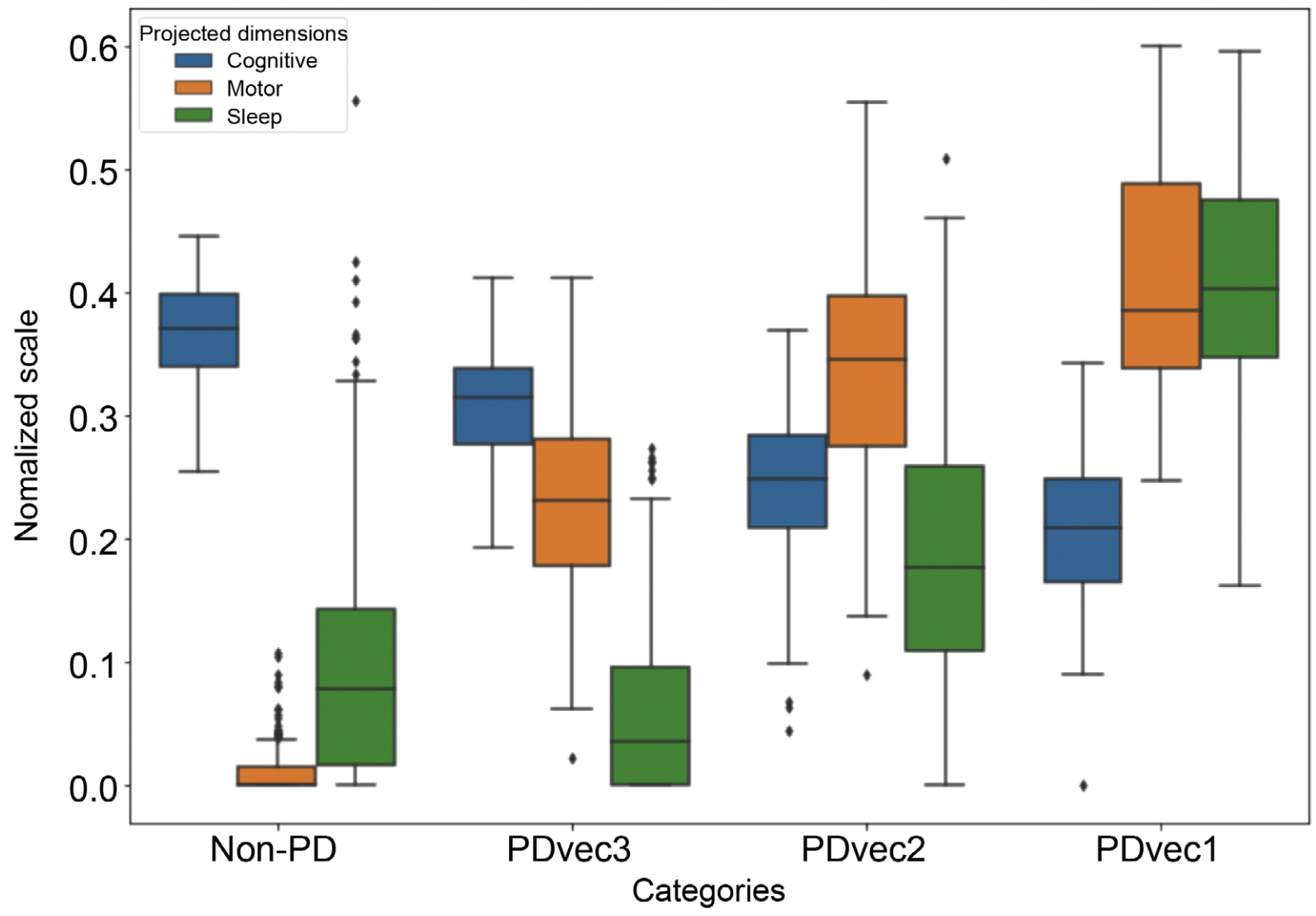

Figure 5. Distribution of projected dimension.

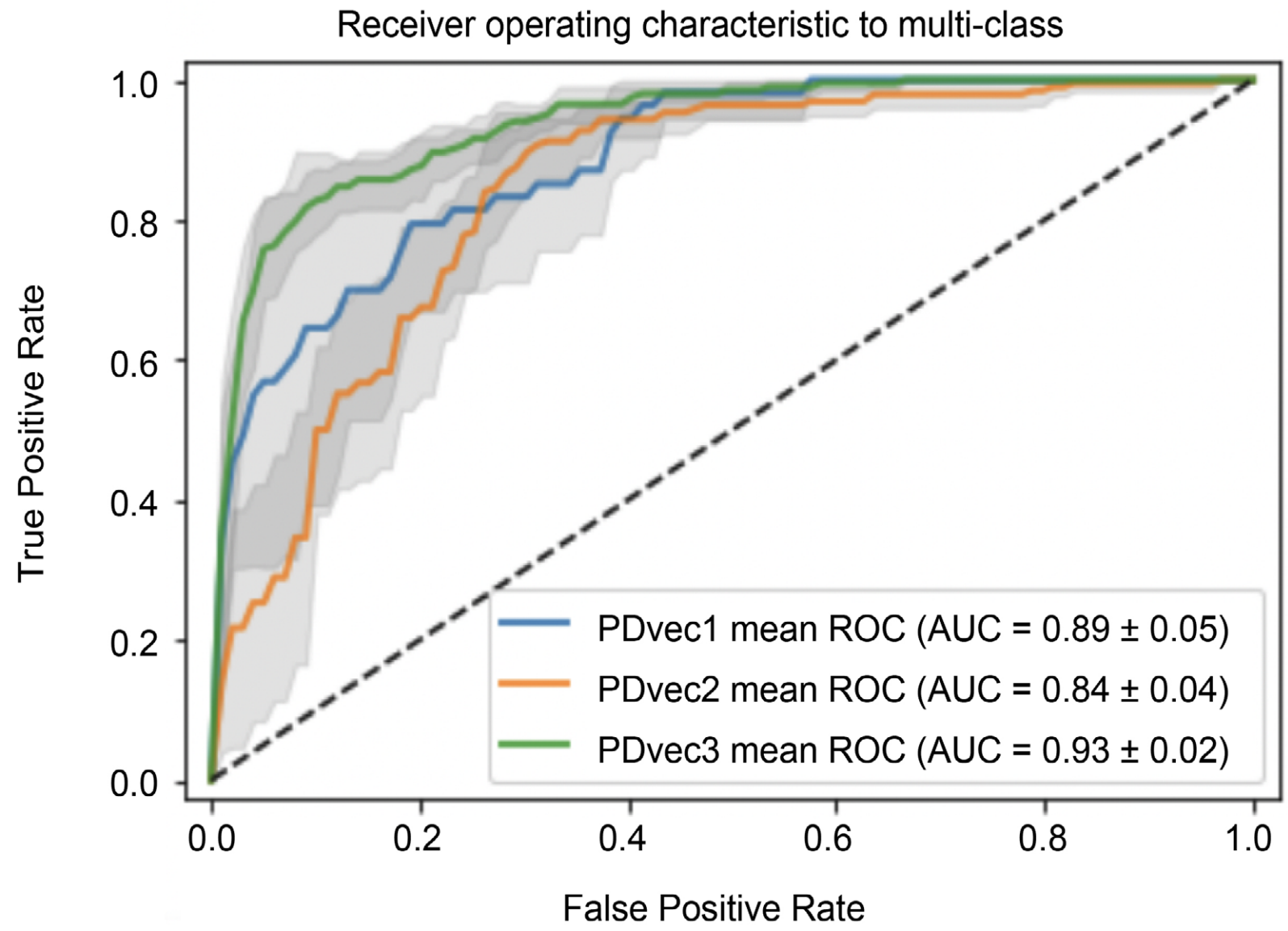

Figure 6. Performance evaluation of parkinson's disease progression model.

We were able to replicate this model and predict results from the PDBP cohort. We only had 120 patients and were able to predict with an AUC of 0.54. Replicated model performed very well on PDVec1 and PDVec3 due to imbalance in classes.

## 4. Discussion

While this work primarily emphasizes the importance of machine learning (specifically the confluence of supervised and unsupervised learning applied to PD), it is conceivable that this method is not always superior to classical statistical techniques. In scenarios where a limited number of variables and highly precise null hypotheses are provided, a significance testing framework may offer superior advantages. There is most certainly a gray area between the Bayesian and Classicist viewpoints here, and certain measurements may not fully capture what they were designed to measure. Within this work, we were able to accurately predict the Parkinson's space of a given PD individual after a set amount of time. However, we were only able to validate this properly on one database PDBP; therefore, the issue is that we need more publically available high-quality data like PPMI and PDBP that promotes further research because any substantive research in this field is a step forward. Most rich clinical data for diseases beyond Parkinson's are not publically available, which poses a challenge for research to be done on these diseases. To uncover potentially valuable subtypes or latent variables, future investigations should strive to integrate unsupervised methodologies. These approaches can help reveal hidden factors or subtypes that contribute to the variability observed in patient characteristics. Additionally, incorporating supervised learning techniques becomes crucial in order to forecast latent scores by utilizing baseline symptoms and to predict future symptoms based on these latent scores.

## 5. Conclusion

In conclusion, we were able to accurately predict trajectories for patients at baseline; however, further research is integral in this field that integrated supervised and unsupervised learning to determine patient trajectories. Overall, the PPMI has a primary goal of comprehending the underlying reasons for which there exists these distinct disease trajectories in differing PD patients. In machine learning applications, it becomes paramount to prioritize the prediction of symptom variability among patients rather than merely distinguishing between Parkinson's disease (PD) patients and healthy individuals. The focus for future research should lie in forecasting the variation in future symptom progression by leveraging the baseline measurements. In other words, the emphasis should shift towards accurately predicting the trajectory of symptoms over time, thus enabling more effective treatment and management strategies. This will allow doctors to determine the severity of their patients' symptoms and therefore recommend better prescriptions that help mitigate their trajectories.

## Conflicts of Interest

The author declares no conflicts of interest.